\documentclass{article}

\usepackage{PRIMEarxiv}

\usepackage[utf8]{inputenc} 
\usepackage[T1]{fontenc}    
\usepackage{hyperref}       
\usepackage{url}            
\usepackage{booktabs}       
\usepackage{amsfonts}       
\usepackage{nicefrac}       
\usepackage{microtype}      
\usepackage{lipsum}
\usepackage{fancyhdr}       
\usepackage{graphicx}       
\graphicspath{{media/}}     

\title{Clinical-Longformer and Clinical-BigBird: Transformers for long clinical sequences}

\author{
  Yikuan Li, Ramsey M. Wehbe, Faraz S. Ahmad, Hanyin Wang, Yuan Luo\thanks{\textit{Corresponding author}}  \\
  Feinberg School of Medicine \\
  Northwestern University \\
  \texttt{\{yikuan.li, ramsey.wehbe, faraz.ahmad, hanyin.wang, yuan.luo\}@northwestern.edu} \\
}

\begin{document}
\maketitle

\begin{abstract}
Transformers-based models, such as BERT, have dramatically improved performance for various natural language processing tasks. The clinical knowledge enriched model, namely ClinicalBERT, also achieved state-of-the-art results when performed on clinical named entity recognition and natural language inference tasks. One of the core limitations of these transformers is the substantial memory consumption due to their full self-attention mechanism. To overcome this, long sequence transformer models, e.g. Longformer and BigBird, were proposed with the idea of sparse attention mechanism to reduce the memory usage from quadratic to the sequence length to a linear scale. These models extended the maximum input sequence length from 512 to 4096, which enhanced the ability of modeling long-term dependency and consequently achieved optimal results in a variety of tasks. Inspired by the success of these long sequence transformer models, we introduce two domain enriched language models, namely Clinical-Longformer and Clinical-BigBird, which are pre-trained from large-scale clinical corpora. We evaluate both pre-trained models using 10 baseline tasks including named entity recognition, question answering, and document classification tasks. The results demonstrate that Clinical-Longformer and Clinical-BigBird consistently and significantly outperform ClinicalBERT as well as other short-sequence transformers in all downstream tasks. We have made our source code available at \url{https://github.com/luoyuanlab/Clinical-Longformer} the pre-trained models available for public download at: \url{https://huggingface.co/yikuan8/Clinical-Longformer}.
\end{abstract}

\section{Introduction}
Transformer-based models have been wildly successful in setting state-of-the-art results to a broad range of natural language processing (NLP) tasks, including question answering, document classification, machine translation, text summarization and others \cite{devlin2018bert, liu2019roberta, brown2020language}. The successes are also replicated in the clinical and biomedical domain via pre-training language models using clinical or biomedical corpora and then transfer them to a variety of clinical or biomedical downstream tasks, e.g. computational phenotyping \cite{rasmy2021med,yao2019traditional}, automatically ICD coding \cite{zhang2020bert}, knowledge graph completion \cite{yao2019kg} and clinical question answering \cite{wen2020adapting}. 

The self-attention mechanism \cite{vaswani2017attention} is one of the most important components that lead to the successes of transformer-based models, which allows each token in the input sequence to independently interact with every other token in the sequence in parallel. However, the memory consumption of self-attention mechanism quadratically enlarges with sequence length, resulting in impracticable training time and easily reaching the memory limits of GPUs. Consequently, transformer-based models that leverage a full self-attention mechanism, such as BERT and RoBERTa, always have an input sequence length limit of 512 tokens. To deal with this limit when modeling long texts using transformer-based models, the input sequence shall be either truncated to the first 512 tokens, or segmented to multiple small chunks of 512 tokens. If the latter method is applied to a text classification task, an aggregation operation will be added to yield the final output. Both segmentation and truncation of the input sequence will ignore the long-term dependencies spanning over 512 tokens and consequently achieve suboptimal results due to information loss. This drawback of self-attention operation will impact language model pre-training and then be amplified to downstream tasks. In clinical NLP, the models that are applied with the transformer-based approaches also encounter this limitation \cite{gao2021limitations}. For example, the discharge summaries in MIMIC-III, which are always used to predict hospital re-admission \cite{huang2019clinicalbert} or mortality \cite{mahbub2022unstructured}, have 1,435 words in average, far exceeding the 512 tokens limits of BERT like models. 
Recently, researchers have been developing some novel variants of transformers particularly for long sequences that reduce the memory usage from quadratic to linear to the sequence length \cite{beltagy2020longformer, zaheer2020big, ainslie2020etc}. The core idea behind these models is to replace the full attention mechanism with the sparse attention mechanism, which is a blend of sliding windows and reduced global attention. These models are able to be fed in up to 4.096 tokens and boost the performance of several question answering as well as text summarization tasks. However, to the best of our knowledge, these transformers for long sequences are rarely discussed in the clinical and biomedical domain. Therefore, we would like to examine the adaptability of these long sequence models using a series of clinical NLP tasks. In particular, we make the following contributions:
\begin{enumerate}
  \item We pre-train and publicly release two transformer models for long clinical sequences, namely Clinical-Longformer and Clinical-BigBird, using large-scale clinical notes. 
  \item We demonstrate that both Clinical-Longformer and Clinical-BigBird significantly improve the performance of a variety of downstream clinical NLP tasks including question answering, named entity recognition and document classification. 
\end{enumerate}

\section{Related Work}
\subsection{Clinical and Biomedical Transformers}
Transformer-based models, especially BERT \cite{devlin2018bert}, can be enriched with clinical and biomedical knowledge through pre-training on large-scale clinical and biomedical corpora. These domain-enriched models, e.g. BioBERT \cite{lee2020biobert} pre-trained on biomedical publications and ClinicalBERT \cite{alsentzer2019publicly} pre-trained on clinical narratives, set the state-of-the-arts when down-stream applied to clinical and biomedical NLP tasks. Inspired by the success of these domain-enriched models, more pre-trained models were released to boost the performance of NLP models when applied to specific clinical scenarios. For example, Smit et al. \cite{smit2020chexbert} proposed CheXbert to annotate thoracic disease findings from radiology reports, which outperformed previous rule-based labelers with statistical significance. The model was pre-trained using the combination of human-annotated and machine-annotated radiology reports. He et al. \cite{he2020infusing} introduced DiseaseBERT, which infused disease knowledge to the BERT model by pre-training on a series of disease description passages that were constructed from Wikipedia and MeSH terms. DiseaseBERT received superior results on consumer health question answering tasks. Zhou et al. \cite{zhou2021cancerbert} developed CancerBERT to extract breast cancer related concepts from clinical notes and pathology reports, which would facilitate the information extraction process. However, all of the aforementioned models were built on the basic BERT architecture, which has a limitation of 512 tokens in the input sequence length. This limitation may result in the information loss of long-term dependencies in the training processes.

\subsection{Transformers for long sequences}
Various attention mechanisms have been proposed to handle the large memory consumption of the attention operations in vanilla Transformer architecture. Transformer-XL \cite{dai2019transformer} segmented a long sequence into multiple small chunks and then learned long-term dependency with a left-to-right segment-level recurrence mechanism. Transformer-XL learns 5.5× dependency compared to the vanilla transformer models, but loses the advantage of bidirectional representation of BERT like models. In another study, Reformer \cite{kitaev2020reformer} applied two techniques to reduce the space complexity of transformer architecture by replacing do-product attention operation with locality-sensitive hashing and sharing the activation function among layers. Reformer was able to process longer sequences at a faster speed and be more memory-efficient. However, no performance improvement can be observed when using Reformer. Almost simultaneously, Longformer \cite{beltagy2020longformer} and BigBird \cite{zaheer2020big} were proposed to drastically alleviate the memory consumption of transformer models from quadratic growth to linear growth to the sequence length, by replacing the pairwise full attention mechanisms with a combination of sliding window attention and global attention mechanisms. They are slightly different in terms of implementation and configuration of the global attention mechanism. Both models support input sequences up to 4.096 tokens long (8× of BERT) and significantly improve performance on long-text question answering and summarization tasks. However, the adaptability of these long sequence transformers to the clinical and biomedical fields, where the text length always exceeds the limits of BERT like models, has been rarely investigated. Levine et al. \cite{simon_levine_2020_4396595} pre-trained longformer models using clinical notes and then fine-tuned the language model on document classification tasks include ICD coding. Our study applied similar approach, while aiming at a more comprehensive comparison to the short sequence models using heterogeneous NLP tasks.

\section{Material and Methods}
In this section, we first introduce the clinical data set we use as the pre-training corpus, followed by the pre-training process of Clinical-Longformer and BigBird. Then, we enumerate the down-stream tasks we used to fine-tune our pre-trained model. We also provide the technical details of pre-training and fine-tuning for reproducing purposes. We have made our source code available at \url{https://github.com/luoyuanlab/Clinical-Longformer} the pre-trained models available for public download at: \url{https://huggingface.co/yikuan8/Clinical-Longformer}.

\subsection{Datasets}
Similarly to \cite{alsentzer2019publicly,huang2019clinicalbert}, we leveraged approximately 2 million clinical notes extracted from the MIMIC-III \cite{johnson2016mimic} dataset, which is the largest publicly available EHR dataset that contains clinical narratives of over 40,000 patients admitted to the intensive care units. We only applied minimal pre-processing steps, including 1) to remove all de-identification placeholders that were generated to protect the PHI (protected health information); 2) to replace all characters other than alphanumericals and punctuation marks; 3) to convert all alphabetical characters to lower cases; and 4) to strip extra white spaces. 
\subsection{Pre-training}
Longformer \cite{beltagy2020longformer} and BigBird \cite{zaheer2020big} are the two most remarkable transformer models that are designed for long input sequences. Both models extend the maximum input sequence length to 4,096 tokens, which is 8× of the conventional transformer-based models by introducing the localized sliding window and global attention mechanisms to reduce the computational expenses of full self-attention mechanisms. The differences between two models are how the global attention is realized and the selection of loss function in fine-tuning. BigBird also contains some random localized attention operations. The performance difference between two models on downstream tasks is minimal. Therefore, we would like to pre-train both models and compare their performance on clinical NLP tasks. We refer readers to the original papers of Longformer \cite{beltagy2020longformer} and BigBird \cite{zaheer2020big} for more technical details.

Byte-level BPE (Byte-Pair-Encoding) \cite{wang2020neural} was applied to tokenize the clinical corpus. We initialized Clinical-Longformer and Clinical-BigBird from the pre-trained weights of the base version of Longformer and the ITC version of BigBird, respectively (although the ETC version performs better, HuggingFace, the largest shared transformers community only provides the implementation and the pre-trained weights of ITC version). Both models were distributed in parallel to 6 32GB Tesla V100 GPUs. FP16 precision was enabled to accelerate training. We pre-trained Clinical-Longformer for 200,000 steps with batch size of 6×3, and Clinical-BigBird for 300,000 steps with batch size of 6×2. The learning rates were 3e-5 for both models. The entire pre-training process took more than 2 weeks for each model.

\subsection{Tasks}
In this study, we adapted the pre-trained Clinical-Longformer and Clinical-BigBird to 9 clinical NLP datasets. These 9 NLP datasets broadly cover various NLP tasks including extractive question answering, named entity recognition, natural language inference, and document classification. We relied on these NLP tasks to validate the performance improvement of our proposed models. The statistics and descriptions of all dataset can be found in Table \ref{tab:1}.

\begin{table}[]
\centering
\caption{Description and statistics of downstream NLP tasks}
\label{tab:1}
\resizebox{\textwidth}{!}{%
\begin{tabular}{@{}lllllll@{}}
\toprule
\textbf{Dataset} & \textbf{Task} & \textbf{Data Source} & \textbf{Domain} & \textbf{Sample Size} & \textbf{Avg. Seq.  Length} & \textbf{Max Seq.  Length} \\ \midrule
MedNLI             & Inference           & MIMIC    & Clinical   & 14,049  & 38.9    & 409    \\
i2b2 2006          & NER                 & i2b2     & Clinical   & 66,034  & 867     & 3,986  \\
i2b2 2010          & NER                 & i2b2     & Clinical   & 43,947  & 1,459.3 & 6,052  \\
i2b2 2012          & NER                 & i2b2     & Clinical   & 13,108  & 793.6   & 2,900  \\
I2b2 2014          & NER                 & i2b2     & Clinical   & 83,466  & 5,133.5 & 14,370 \\
emrQA-Medication   & QA                  & i2b2     & Clinical   & 255,908 & 1,880.3 & 6,109  \\
emrQA-Relation     & QA                  & i2b2     & Clinical   & 141,243 & 1,460.3  & 6,050  \\
emrQA-HeartDisease & QA                  & i2b2     & Clinical   & 30,731  & 5,292.7 & 14,060 \\
OHSUMed            & Multiclass Classif. & MEDLINE  & Biomedical & 4,056   & 301.1   & 861    \\
OpenI              & Multilabel Classif. & IndianaU & Clinical   & 3,684   & 69.7    & 294    \\
MIMIC-AKI          & Binary Classif.     & MIMIC    & Clinical   & 16,536  & 1463.1  & 20,857 \\ \bottomrule
\end{tabular}%
}
\end{table}

\subsubsection{Question Answering}
Question answering (QA) is an NLP discipline, which targets to automatically answer questions raised from human languages \cite{cimiano2014ontology}. In the clinical context, QA system answers physicians' questions by understanding the clinical narratives extracted from electronic health record systems to support decision making. emrQA \cite{pampari2018emrqa} is the most frequently used benchmark dataset in clinical QA, which contains more than 400,000 question-answer pairs semi-automatically generated from past i2b2 challenges. emrQA falls into the category of extractive question answering that aims at identifying the answer spans from reference contexts instead of generating answers in a word-by-word fashion. Researchers have attempted to solve emrQA tasks by using word embedding models \cite{yue2020clinical}, conditional random fields (CRFs) \cite{kang2020neural} and transformer-based models \cite{soni2020evaluation}, among which transformer-based models defeated their competitors in terms of performance. In our experiments, we investigated the performance of our pre-trained models using the three largest emrQA subsets: \textit{Medication}, \textit{Relation} and \textit{Heart Disease}.  We evaluated QA performance with two commonly used metrics: exact match (EM) and F1-score. Exact match represents that the predicted spans and categories exactly and strictly match with the ground-truth annotations. F1-score is a looser metric derived from token-level precision and recall, which aims at measuring the overlap between the predictions and the targets. We generated train-dev-test splits by following the instruction of \cite{yue2020clinical}, where the training set of relation and medication subsets were randomly under-sampled to reduce training time. Based on their experience, the performance would not be compromised after under-sampling. 
\subsubsection{Named Entity Recognition}
Named-entity recognition is a token-level classification task that seeks to identify the named entities and classify them to predefined categories. This genre of NLP tasks has broad applications in the clinical and biomedical domain. e.g. de-identification of PHI, and medical concept extraction from clinical notes. Prior studies have shown that Transformer-based models \cite{lee2020biobert} significantly outperformed the models built on pre-trained static word embedding \cite{yoon2019collabonet} or LSTM networks \cite{wang2019cross}. We examined our pre-trained models using four data challenges of: 1) i2b2 2006 \cite{uzuner2007evaluating}, to de-identify PHI from medical discharge notes; 2) i2b2 2010 \cite{uzuner20112010}, to extract and annotate medical concepts from patient reports; 3) i2b2 2012 \cite{sun2013evaluating}, to identify both clinical concepts and events relevant to the patient's clinical timeline from discharge summaries; and 4) i2b2 2014 \cite{stubbs2015annotating}, to identify PHI information from longitudinal clinical narratives. We followed the processing steps of\cite{alsentzer2019publicly}, which converted the raw data from all four tasks to IOB (inside–outside–beginning) tagging format proposed by Ramshaw et al, and then created train-dev-test splits. \cite{ramshaw1999text}. Based on the new tagging format, we evaluated the model performance with F1-score similarly to QA tasks. 
\subsubsection{Document Classification}
Document classification is one of the most common NLP tasks, where we assign a sentence or document to one or more classes or categories. In the clinical domain, researchers pay more attention to identifying the diagnosis, or to predicting the onset of a certain disease from clinical notes. We applied the following four document classification data set to evaluate the pre-trained models from different perspectives. \\
\newline
\textbf{MIMIC-AKI \cite{li2018early,sun2019early}} MIMIC-AKI is a binary classification task, where we would like to predict the possibility of AKI (acute kidney injury) for critically ill patients using the clinical notes within the first 24 hours following intensive care unit (ICU) admission. We evaluated the model performance using AUC and F1 score. \\
\\
\textbf{OpenI \cite{demner2012design}} OpenI is a publicly available chest x-ray (CXR) dataset collected by Indiana University. The dataset provides around 4,000 radiology reports and their associated human annotated MeSH terms. Given the small sample size, we only used openI as the testing set. The pre-trained language models were fine-tuned using MIMIC-CXR \cite{johnson2019mimic}, another publicly available chest x-ray dataset. Unlike openI, the ground-truth labels for MIMIC-CXR were automatically generated using NLP approaches. The annotated findings overlap between two CXR data sources are the diagnosis of: \textit{Cardiomegaly}, \textit{Edema}, \textit{Consolidation}, \textit{Pneumonia}, \textit{Atelectasis}, \textit{Pneumothorax} and \textit{Pleural Effusion}. In experiments, we would like to retrieve the presences of all possible thoracic findings from CXR reports. This task can be considered as a multi-label classification task. We would report the sample number weighted average of AUC as proposed in \cite{wang2018tienet}.\\
\\
\textbf{MedNLI \cite{romanov2018lessons}} Natural language inference (NLI) is the task of determining the relationship between sentence pairs. MedNLI is a collection of sentence pairs extracted from MIMIC-III \cite{johnson2016mimic} and annotated by two board-certified radiologists. The relationship between the premise sentence and the hypothesis sentence could be entailment, contradiction or neutral. Transformer-based models process NLI tasks also as document classification by merging the sentence pair together and placing a delimiter token right after the end of the first sentence. We followed the original splits as \cite{romanov2018lessons} and used accuracy to evaluate the performance.\\
\\
\textbf{OHSUMed \cite{hersh1994ohsumed}} OHSUMed is a collection of journal titles and/or abstracts extracted from MEDLINE, which is maintained and published by the National Library of Medicine. Unlike other data sources aforementioned, this dataset is the only one that the free-text raw data is not obtained from clinical notes but academic journals. Following \cite{yao2019graph}, only those cardiovascular diseases abstracts that are associated with one and only one of the 23 disease categories are included in this study. We also used accuracy to evaluate this multi-class classification task.
\subsection{Baseline Models}
Both Clinical-Longformer and Clinical-BigBird were compared with the pioneering BERT, ClinicalBERT and BioBERT models. BERT \cite{devlin2018bert} model is the first-of-its-kind transformer architecture that achieved state-of-the-art results in eleven NLP tasks. Both masked language modeling and next sentence prediction were used to learn contextualized word representation from BooksCorpus and English Wikipedia in the pre-training stage. BioBERT \cite{lee2020biobert} is the first biomedical domain specific BERT variant that was pre-trained from PubMed abstracts and PMC full-text articles. The weights of BioBERT were initialized from BERT. BioBERT yielded optimal performance in biomedical QA, NER and relation extraction tasks. ClinicalBERT \cite{alsentzer2019publicly}, initialized from BioBERT, was further pre-trained using clinical notes that were also extracted from MIMIC-III. ClinicalBERT boosted the performance for MedNLI and four i2b2 NER tasks that were also included in our study. Both BioBERT and ClinicalBERT discarded the next sentence prediction task and only selected the masked language modeling as the pre-training strategy. 
\subsection{Experimental setup}
For the token-level classification, including QA and NER, a classification head was added to the output of each token obtained from the transformer-based architecture. The maximum sequence length was selected to be 3,072 for Clinical-Longformer and Clinical-BigBird, and 384 for all the other three baseline models. For the sequence-level classification, the predicted outcomes were yielded from the [CLS] token added to the beginning of each sentence or document. Given that some clinical notes are extremely long, which may even exceed the length limits of all models, we first truncated each document/sentence/sentence-pair to 4096 tokens, which  met the length limits of Clinical-Longformer and Clinical-BigBird. The document/sentence/sentence-pair was further segmented to snippets of 512 tokens in order to accommodate for the length requirement of BERT/BioBERT/ClinicalBERT. A pooling layer was applied to aggregate the output from short snippets. We conducted our experiments using four 32GB gpus. The batch size during training was 4×4 for Clinical-Longformer, 4×3 for Clinical-BigBird, and 4×16 for all other models. Batch sizes were doubled when evaluating or testing. Half precision was applied to both Clinical-Longformer and Clinical-BigBird. The only hyper-parameter we tuned was learning rate. We tried {1e-5, 2e-5 and 5e-5} for the experiments of each model on each task. We fine-tuned 6 epochs for each set-up. The best-performing model parameters were determined by the performance on the developing split. The experiments were implemented with python 3.8.0, pytorch 1.9.0 and Transformer 4.9.0. 

\section{Results and Discussion}
Full results for QA, NER and classification tasks are presented in Table \ref{tab:2}, \ref{tab:3}, \ref{tab:4}, respectively. In question answering tasks,  both Clinical-Longformer and Clinical-BigBird outperformed the pioneering short-sequence transformer models ~ 2 percents across all three emrQA subsets when evaluated by F1 score. When considering the stricter EM metric, Clinical-Longformer and Clinical-BigBird improved ~ 5 percent on the relations subset, but yielded similar results to ClinicalBERT in the other two subsets. In NER tasks, Clinical-Longformer consistently led the short-text transformers by more than 2 percents in all 4 i2b2 dataset. Clinical-BigBird also performed better than ClinicalBERT and BioBERT in every single NER experiments. In document classification tasks, our two long-sequence transformers achieved better results compared to prior models on OpenI, MIMIC-AKI and medNLI, which were all generated from clinical data sources. BioBERT performed slightly better than our models in OHSUMed dataset. We argue that OHSUMed is the only task that falls under the scope of biomedical data mining, making it align better with BioBERT which was pre-trained from biomedical publications. 

We observed that Clinical-Longformer and Clinical-BigBird not only improved the performance of long sequences tasks, but also short sequences. The maximum sequences of MedNLI and OpenI are smaller than 512 tokens, which can be entirely fed into the BERT like models. However, the long sequences models still achieved better results. We attributed these improvements to the pre-training stages of Clinical-Longformer and Clinical-BigBird, where the language models can learn more long-term dependencies by extending the limit of sequence length. In addition, we found that more significant improvement can be achieved when applying our two models to the dataset with longer sequences. For example, the performance improvement on i2b2 2014, which has the largest averaged sequence length in all 4 NER tasks, is almost twice of the other three subsets. Likewise, Clinical-Longformer more strongly improved the F1 score of the heart disease subset from emrQA. This helps us confirm that Clinical-Longformer and Clinical-BigBird are also able to model long-term dependency in the downstream tasks. We also noticed that Clinical-Longformer consistently yielded better results when compared to Clinical-BigBird. Given that Clinical-BigBird also requires more fine-tuning time, we recommend readers to apply our Clinical-Longformer checkpoint to their own tasks when they have limited resources. 

Our study has several limitations, each of which can lead to further investigations. First, we only applied Longformer and BigBird to large scale clinical corpora. In the next step, we plan to release more pre-trained models for long sequences enriched with other biomedical corpora, e.g. PubMed and PMC publications. Second, another approach developed to address the memory problem of long sequences is to simplify or compress the transformer architecture. In our future work, we will compare this genre of transformers, e.g. TinyBERT to our current long sequence models. Last, we did not integrate Clinical-Longformer or Clinical-BigBird into an encoder-decoder framework due to the memory limits of our GPU cards. Therefore, the experiment on text generation or summarization is not included in this study. We will definitely incorporate those tasks into our future study, once we acquire more computational resources.

\begin{table}[]
\centering
\caption{The performance of transformer-based pre-trained models on question answering tasks.}
\label{tab:2}
\resizebox{0.85\textwidth}{!}{%
\begin{tabular}{@{}lcccccc@{}}
\toprule
\textbf{Pre-trained Models} &
  \multicolumn{2}{l}{\textbf{emrQA - Medication}} &
  \multicolumn{2}{l}{\textbf{emrQA - Relation}} &
  \multicolumn{2}{l}{\textbf{emrQA- Heart Disease}} \\ \cmidrule(l){2-7} 
\multicolumn{1}{r}{\textit{metrics}} & \textit{EM}    & \textit{F1}    & \textit{EM}    & \textit{F1}    & \textit{EM}    & \textit{F1}    \\ \midrule
BERT                                 & 0.240          & 0.675          & 0.833          & 0.924          & 0.650          & 0.698          \\
BioBERT                              & 0.247          & 0.700          & 0.836          & 0.926          & 0.647          & 0.702          \\
ClinicalBERT                         & 0.297          & 0.698          & 0.849          & 0.929          & 0.666          & 0.711          \\
Clinical-Longformer                  & \textbf{0.302} & \textbf{0.716} & \textbf{0.911} & \textbf{0.948} & \textbf{0.698} & \textbf{0.734} \\
Clinical-BigBird                     & 0.300          & 0.715          & 0.898          & 0.944          & 0.664          & 0.711          \\ \bottomrule
\end{tabular}%
}
\end{table}

\begin{table}[]
\centering
\caption{The performance of transformer-based pre-trained models on named entity recognition tasks.}
\label{tab:3}
\resizebox{0.65\textwidth}{!}{%
\begin{tabular}{@{}lcccc@{}}
\toprule
\textbf{Pre-trained Models} &
  \multicolumn{1}{l}{\textbf{i2b2 2006}} &
  \multicolumn{1}{l}{\textbf{i2b2 2010}} &
  \multicolumn{1}{l}{\textbf{i2b2 2012}} &
  \multicolumn{1}{l}{\textbf{i2b2 2014}} \\ \cmidrule(l){2-5} 
\multicolumn{1}{r}{\textit{metrics}} & \textit{F1}    & \textit{F1}    & \textit{F1}    & \textit{F1}    \\ \midrule
BERT                                 & 0.939          & 0.835          & 0.759          & 0.928          \\
BioBERT                              & 0.948          & 0.865          & 0.789          & 0.930          \\
ClinicalBERT                         & 0.951          & 0.861          & 0.773          & 0.929          \\
Clinical-Longformer                  & \textbf{0.974} & \textbf{0.887} & \textbf{0.800} & \textbf{0.961} \\
Clinical-Big Bird                    & 0.967          & 0.872          & 0.787          & 0.952          \\ \bottomrule
\end{tabular}%
}
\end{table}

\begin{table}[]
\centering
\caption{The performance of transformer-based pre-trained models on document classification tasks.}
\label{tab:4}
\resizebox{0.66\textwidth}{!}{%
\begin{tabular}{@{}lccccl@{}}
\toprule
\textbf{Pre-trained Models} &
  \multicolumn{1}{l}{\textbf{OHSUMed}} &
  \multicolumn{1}{l}{\textbf{OpenI}} &
  \multicolumn{2}{l}{\textbf{MIMIC-AKI}} &
  \textbf{medNLI} \\ \cmidrule(l){2-6} 
\multicolumn{1}{r}{\textit{metrics}} &
  \textit{Accuracy} &
  \textit{AUC} &
  \textit{AUC} &
  \textit{F1} &
  Accuracy \\ \midrule
BERT              & 0.717 & 0.952 & 0.514 & 0.293 & 0.776 \\
BioBERT           & \textbf{0.771} & 0.954 & 0.534 & 0.324 & 0.808 \\
ClinicalBERT      & 0.741 & 0.967 & 0.738 & 0.444 & 0.812 \\
Clinical-Longformer &
  0.766 &
  \textbf{0.977} &
  \textbf{0.762} &
  \textbf{0.484} &
  \textbf{0.842} \\
Clinical-Big Bird & 0.752 & 0.972 & 0.755 & 0.480 & 0.827 \\ \bottomrule
\end{tabular}%
}
\end{table}

\section{Conclusion}
In this study, we introduced Clinical-Longformer and Clinical-BigBird, two pre-trained language models that are suitable for long sequence clinical NLP tasks. We compared these two models with the pioneering short sequence transformer based models, e.g. ClinicalBERT, in named entity recognition, question answering, and document classification tasks. Results demonstrated that Clinical-Longformer and Clinical-BigBird achieved better results in both long sequence and short sequence benchmark dataset. Future studies will be investigated on other models, for example TinyBERT, that are developed to solve the memory consumption of long sequences as well as the generalizability of our proposed models to clinical text generation and summarization.

\section*{Acknowledgments}
This was was supported in part by NIH grants U01TR003528 and R01LM013337.

\bibliographystyle{unsrt}  
\bibliography{longformer}

\end{document}